\documentclass[runningheads]{llncs}

\usepackage{times} % assumes new font selection scheme installed
\usepackage{amsmath} % assumes amsmath package installed
\usepackage{amssymb}  % assumes amsmath package installed
\usepackage[linesnumbered,ruled]{algorithm2e}
\usepackage{subfigure}
\usepackage{graphicx}

\usepackage{multirow}
\usepackage[inline]{enumitem}
\usepackage[export]{adjustbox}
\usepackage[english]{babel}
\usepackage[utf8]{inputenc}
\let\oldnl\nl% Store \nl in \oldnl
\newcommand{\nonl}{\renewcommand{\nl}{\let\nl\oldnl}}
\usepackage[noend]{algpseudocode}
\usepackage{caption}
\usepackage{mathtools}

\usepackage{xcolor}
\usepackage{soul}
\usepackage{bm}
\usepackage{array}
\newcolumntype{M}[1]{>{\centering\arraybackslash}m{#1}}
\usepackage[utf8]{inputenc}

\begin{document}
\title{Loosely Coupled Payload Transport System with Robot Replacement}

\author{Pulkit Verma \and
Rahul Tallamraju \and
Abhay Rawat \and
Subhasis Chand \and
Kamalakar Karlapalem}
\authorrunning{P. Verma et al.}
\institute{International Institute of Information Technology, Hyderabad, India\\
% \email{kamal@iiit.ac.in}\\
% \url{https://www.iiit.ac.in/people/faculty/kamal/} \\
\email{\{pulkit.verma@research., rahul.t@research., abhay.rawat@research., subhasis.chand@research., kamal@\}iiit.ac.in}}
\maketitle              % typeset the header of the contribution
%
% We propose a novel concept which ensures that the multi-robot system lasts much longer than the battery life of an individual robot.

\begin{abstract}
In this work, we present an algorithm for robot replacement to increase the operational time of a multi-robot payload transport system. Our system comprises a group of non-holonomic wheeled mobile robots traversing on a known trajectory. We design a multi-robot system with loosely coupled robots that ensures the system lasts much longer than the battery life of an individual robot. A system level optimization is presented, to decide on the operational state (charging or discharging) of each robot in the system. The charging state implies that the robot is not in a formation and is kept on charge whereas the discharging state implies that the robot is a part of the formation. Robot battery recharge hubs are present along the trajectory. Robots in the formation can be replaced at these hub locations with charged robots using a replacement mechanism. We showcase the efficacy of the proposed scheduling framework through simulations and experiments with real robots.

\end{abstract}
\scriptsize
\subsubsection{Video Link: https://youtu.be/-6ivGT3dOQw}
\normalsize
\section{Introduction}
Coordination between multiple robotic agents to collectively perform tasks such as payload transportation \cite{transport1} \cite{transport2}, search and rescue \cite{snr1} \cite{snr2} and area exploration \cite{ae1} \cite{ae2}, have been a field of interest. Advantages of using multiple robots over a single robot have been well established in certain application domains \cite{cao} \cite{parker} \cite{Yang}. Battery life of a robot is a crucial aspect \cite{social_management} \cite{drones_battery} \cite{egerstedt_battery} in highly coordinated tasks and applications such as payload transportation. Complete battery discharge of a single robot during payload transportation can render the task incomplete. \textit{In this paper, we focus on robots replacement in a multi-robot payload transport system such that the operational time of the system is well beyond the battery life of an individual robot. Robot replacement ensures that the high-level task of payload transportation remains less dependent on the robot's battery life, we present a battery discharge aware robot replacement mechanism}.  

% Hubs contain fully or partially charged robots, which are used for replacement of low battery robots in formation using our algorithm.
%Fully or partially charged robots are present in recharge hubs which are used to replace low battery robots in formation. This replacement is carried out using our presented algorithm.
\begin{figure}
	\includegraphics[height=0.90\textheight, width=\textwidth]{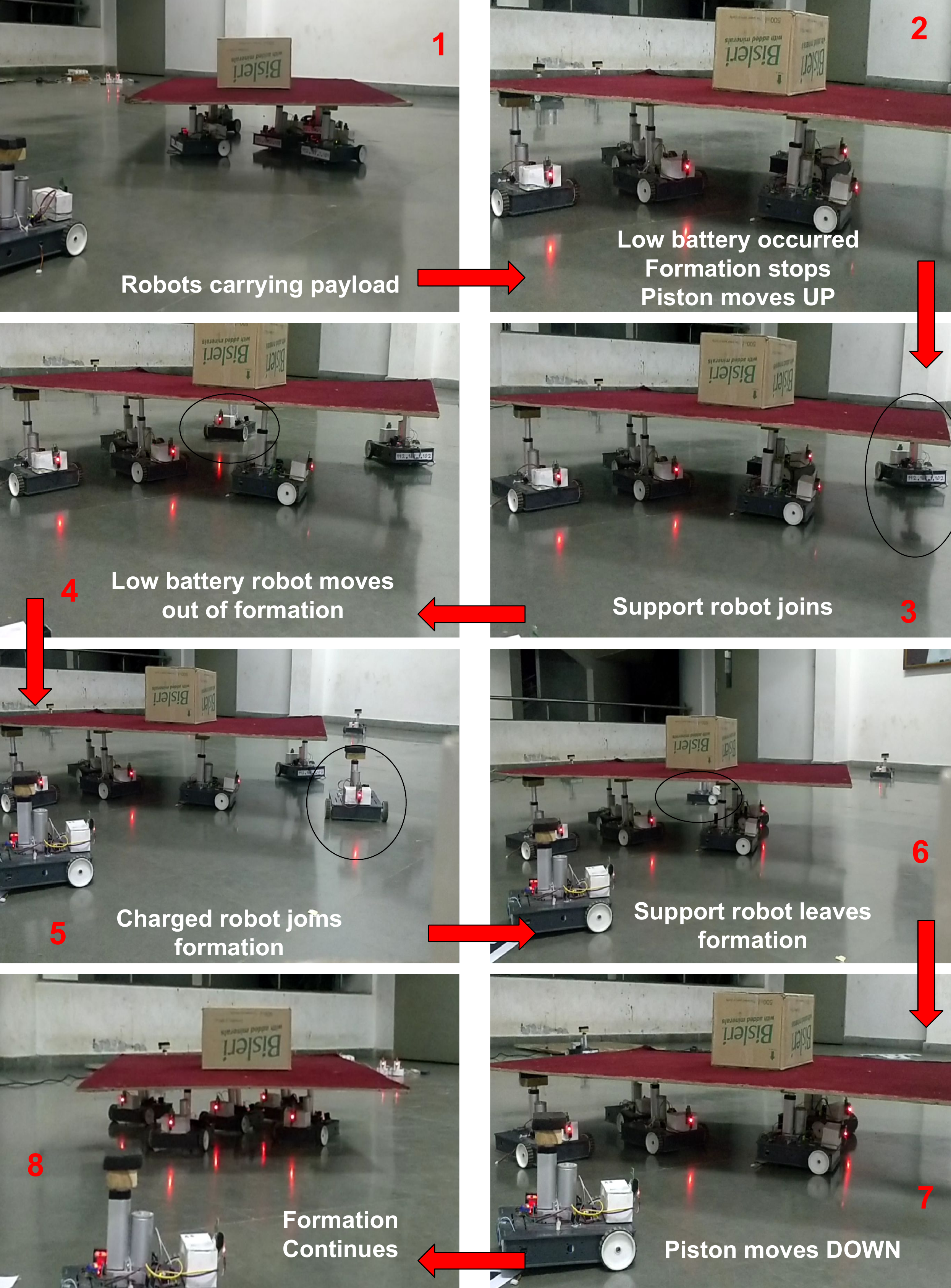}
	\caption{Robot Replacement Process (for more details, see video at https://youtu.be/-6ivGT3dOQw)}
	\label{hardware}
\end{figure}
% {enumerate*}[label=\itshape\alph*\upshape)]
In our work, a formation of robots transports payloads.
The robots in a formation refer to the robots that are contributing to transporting the payload while the other robots are distributed among the charging stations (or recharge hubs). Recharge hubs are present alongside the trajectory to carry out replacements. The formation carrying the payload traverses along a trajectory used by a leader robot. Other robots in the formation derive their trajectories using a decentralized control law \cite{formation} with respect to the designated leader robot. Fully or partially charged robots are present in recharge hubs which are used to replace low battery robots in the formation using our algorithm. \par 
An optimization problem is formulated to \begin {enumerate*}[label=(a),font=\itshape]
    \item extend the operational time of the complete system and
\end{enumerate*} 
\begin {enumerate*}[label=(b),font=\itshape]
    \item reduce the number of replacements while traversing the trajectory,
\end{enumerate*} adhering to the system constraints. Fig.\ref{hardware} depicts the series of steps that are followed to carry out a replacement in a formation.\par
% explains the sequence of steps that take place while running the system. .We perform simulation as well as hardware experiments to show that our solution ensures energy aware system operation and increases the operational duration of the multi-robot payload transport system.
We also demonstrate the hardware results (in Section \ref{hardware_implementation}) of the presented payload transport system with five robots moving in a formation carrying a payload.
% We demonstrate the hardware results of a payload transport system, (also discussed in Section \ref{hardware_implementation}) where robots move in a formation, carrying a payload on them. 
A circular trajectory is considered with two recharge hubs on its periphery containing one robot at each hub. Replacements are carried out at any of these hubs as dictated by our algorithm. Robots are equipped with current and voltage measuring circuit which is used to calculate the battery consumption of each robot. The consumption data is sent to an optimization solver running on a server. The optimization provides a binary solution vector, where $1$ represents robot in a formation and $0$ represents robot at recharge hub. We identify the low battery robot(s) and charged robot(s) by using the solution vectors generated at the current and previous recharge hub respectively (refer to \ref{optimization_formulation}). The replacement of the low battery robots is carried out with charged robots present at the hub without any human intervention. In order to automate robot replacements and maintain payload stability during robot replacements, we utilize \textit{Support Robots}. These are typically present at recharge hubs and are not considered while modeling the optimization. \par
% Here, we introduce a concept of \textit{Support Robots} that are not considered while modeling the optimization problem, but helps the formation maintaining the balance of the payload at the time replacement. The \textit{Support Robot}  are used only during the replacement of robots whenever required.

% will join the formation at the time of replacement on a temporary basis to provide support to the payload so that the payload does not fall. Once the replacement process is completed, the support robot moves out of the formation. 
% The support robots are also present at the recharge hubs or nearby. \par
% but are not considered in optimization for replacement. One support robot is also considered while performing experiments. 
% A payload of six kgs is considered for testing purpose. 
For physical validation, we consider a payload weighing six kgs. From practical implementation, we observed that the system has an operating time of $40$ minutes without any payload and $25$ minutes with the payload. Using our proposed robot replacement algorithm, we observe that the operational duration of the system increases by about $20$ minutes i.e.\ the operating time is increased to nearly $1$ hour with a payload on them. All the robots in a formation are left with minimal battery after an hour and hence replacement cannot be made as the robots available on the hubs are limited in our case. Increasing the number of robots at the hubs will result in a further increase in the operational time of the system. Note that there is no limit for carrying out the number of replacements i.e.\ we can make multiple replacements at any hub even if more than one low battery robot is present in the formation, provided enough robots are present at the recharge hub for replacement. However, the multiple replacements are carried out one at a time to avoid instability of the payload. To avoid a long wait time and to showcase the replacement process, the battery threshold for replacement is set to $11.5$ volts (having a maximum voltage of $12$ volts). If there are no charged robots present at the hub for the replacement(s) of a low battery robot(s), the formation will stop and wait for the replacement robot to be available\footnote{Video at: https://youtu.be/-6ivGT3dOQw}. \par
Some assumptions are made to showcase the efficacy of presented solution on real robots:

% \subsubsection*{Assumptions}
% The idea presented in this paper is implemented on real robots for demonstrating the efficacy of our system. To make the system practically implementable we have considered some assumptions as given below. 

% The proposed idea is implemented on the real robots demonstrating the significance and efficacy of our system. The real environment poses several constraints on the simulated environment. We present these constraints by considering few assumptions, given as
\begin{itemize}
    % \item Robots are non-holonomic wheeled mobile robots.
    \item The robots are moving on flat terrain. Inclined planes are not considered in this paper.
    \item The battery in each robot is fault free and has an error free charging and discharging cycle.
    \item Only low battery faults are considered at this time and no other robot faults like wheel failure, piston failure, etc are considered in this paper.
    \item As robot replacements typically happen near recharge hubs, we assume that the support robots are always available during robot replacements. 
    % \item As we are using the concept of support robots, the payload remains balanced and does not fall when robot replacement is carried out.
    % \item We use more number of robots in a formation than required to carry the payload. These extra robots are called \textit{Redundant Robots}.
    % \item The total number of robots (excluding robots in formation and support robots) are distributed among the recharge hubs. If there are no robots available at the recharge hubs to carry out replacements, the system will wait for the replacement to be available. 
    % \item All robots can have different initial battery charge.
    
    % What if battery of support robot goes down, if one robots is always available at each hub for replacement
    
    % add concept of support robots
        % \item Robots are present at the recharge hub for replacement and providing support to the payload. If there are no robots available at the recharge hub for carrying out replacements, the system stops. 
\end{itemize}

\section{Related Work}
\label{related_work}
Cooperative payload transport and manipulation using multiple robots has been studied before \cite{transport2} \cite{transport1} \cite{pushing1}. However, to the best of our knowledge, none of the approaches consider maximizing operational time of a loosely coupled formation of robots for payload transportation. Therefore, we review literature related to different aspects of our work. Battery aware approaches to plan and control, use battery models as constraints to improve the run time of the system. \cite{egerstedt_battery} \cite{rendezvous2} solves the multi-agent rendezvous problem using constraints on robot kinematics and battery. Time of flight of drones is maximized by using mobile ground recharge stations to charge the drones during real-time operations \cite{drones_battery}. While we consider maximization of operational time in our work, we plan for the formation of mobile robots with task constraints to transport a common payload. In \cite{b_aware_explore}, the robots are used for area exploration in an unknown environment and recharge them on charging points present at a fixed location. The author considers the battery parameter for charging and discharging but does not deal with the replacement of robots. Battery is shared among robots \cite{truly} \cite{potential} \cite{tanker}, where each robot has multiple battery packs. The batteries can optionally be shared on-the-go with other robots in the vicinity, which are running low on battery charge. Even though the authors solve the issues of increasing system lifetime by increasing the number of batteries, but the approach, however, involves extra effort to carry additional battery weight. In our approach, we use a single battery robot and recharge them when running low on charge. However, this can be extended to multiple batteries on a single robot.\par
The concept of working robots in the home area and foraging area is discussed in \cite{optim}, where the robots in the home area move to perform a task and robots in a foraging area search for known or unknown power stations. As the power of the robots in the home area is reduced to a certain threshold, the forage robots help them to find the nearest power station. Though the author deals with recharging the robots and increasing the battery life but does not deal with robot replacement in the formation of robots. Time and energy constrained schedules have been generated for multiple robotic manipulators using a mixed integer nonlinear program \cite{battery_optim}.  Parameters like battery models and component power consumption were considered \cite{social_management} for effective long-term power management in a socially constrained multi-robot system. Failure resilient multi-agent system is considered in \cite{rahul} where a robot with multiple wheels is used to carry a payload. The author deals with minimizing the energy of the system and maximizing the traveled distance by switching on and off the motors (agents) such that the robot trajectory remain unaffected. While the aforementioned works study battery constrained planners and schedulers, they do not deal with the task of formation, payload and battery constrained multi-robot systems.

\subsection*{Contribution and Organization}
In our work, the robots are given a task of transporting a payload. In this paper, we present an algorithm for task constrained optimal robot replacement to extend the operational time of a multi-robot system. We formulate an optimization problem to determine the set of discharged robots, which have to be replaced by charged robots from a recharge hub. Our optimization is a quadratic program which is constrained by (a) the number of robots required to transport the payload, (b) the battery levels of each robot in the formation and (c) battery levels of robots at recharge hub stations. The nature of the modeled optimization problem is quadratic as the voltage of the robot reduces quadratically with the number of time steps. (see equation \ref{discharge_eqn}). 
% The robot voltage at a few time steps (or hubs) ahead has a square relation with the input voltage at the current hub and therefore the nature of the modeled optimization problem is quadratic. 
We discuss the battery discharge of a Li-Ion battery and consider kinodynamics of the non-holonomic differential drive robot to simulate and implement our algorithm. 
%We additionally present hardware results to show the effectiveness of the proposed algorithm in maximizing total system running time.  -- REMOVED BY RAHUL
The replacement of the low battery robot with a charged robot is presented using our algorithm. However, to the best of our knowledge, no previous work has been done on extending the operational time of a multi-robot system for the task of payload transportation (considering non-holonomic wheeled mobile robots) using robot replacement.  \par
%The idea presented in the paper is novel and has not yet been implemented to the best of our knowledge.
We discuss the kinematics, dynamics and battery model of a differential drive robot in Section \ref{Modeling}. We discuss about modeling an optimization problem to find the charged and low battery robots at a hub and in the formation respectively in Section \ref{Scheduling}. The section also describes a replacement strategy to replace low battery robots with charged robots. To make our solution more concrete, we display simulation and experimental results in Section \ref{simulations}. Conclusion and future scope is discussed in Section \ref{conclusion} 
    % Related work in this domain is discussed in Section \ref{related_work}, followed by conclusion and future scope in Section \ref{conclusion}.

\section{Modeling and Control}\label{Modeling}
% This section deals with discussion on various types of models of a differential drive robot. 
This section deals with a discussion on kinematics, dynamics, and frictional model, followed by a battery model of each robot and an approach for making multiple robots to move in a loosely coupled formation.
% In this section we discuss the robot kinematic, dynamic and electrical models. Additionally, we present the decentralized formation control law.
\subsection{Kinematics Model}
%\begin{figure}[h]
%	\includegraphics[scale = 0.2, center]{Diff_drive}
%	\caption{A non-holonomic differential drive robot}
%	\label{Diff_drive}
%\end{figure}

A nonholonomic robot is considered, with generalized coordinates $\bm{q} = [x, y, \theta, \phi_r, \phi_l]^T$, where $(x, y)$ is the position in the inertial frame and  $\theta$ is the yaw angle,  $\phi_r$ and $\phi_l$ represent the angular positions of the left and right wheel. The evolution of generalized coordinates with is given by the following equation.
\begin{eqnarray}
\bm{\dot{q}} &=& \bm {S(q)} \bm{\nu} \nonumber \\
\text{where, } \bm{\nu} &=& \begin{bmatrix} 
v && w \\ 	
\end{bmatrix}^T
\end{eqnarray}
and $\bm{S(q)}$ spans the null space of the non-holonomic constraint matrix of the differential drive robot. $v$ and $\omega$ are robot's pseudo linear and angular velocity control inputs. 

\subsection{Dynamics Model}
The dynamics equation for the considered differential drive robot are derived using the Euler-Lagrange formulation \cite{Ramosg}.
\begin{equation} \label{dynamics}
\nonumber \bm{M}(\bm{q})\bm{\dot{\phi}} + \bm{m}(\bm{q}, \bm{\phi}) = \bm{G}^{T}(\bm{q})\bm{S}(\bm{q})\bm{\tau}
\end{equation}
\begin{equation}
\nonumber \bm{M}(q) = \bm{G}^{T}(\bm{q})\bm{B}(\bm{q})\bm{G}(\bm{q})
\end{equation}
\begin{equation}
\bm{m}(\bm{q}, \bm{\phi}) = \bm{G}^{T}(\bm{q})\bm{B}(\bm{q})\bm{\dot{G}}(\bm{q})\bm{\phi} + \bm{G}^{T}(\bm{q}) \bm{C}(\bm{q}, \bm{\dot{q}}) 
\end{equation}
$\bm{G}(\bm{q})$ is the transformation matrix, $\bm{B}(\bm{q})$ is a symmetric positive definite inertia matrix, $\bm{C}(\bm{q}, \bm{\dot{q}})$ is the Coriolis matrix. Also, $\bm{\tau} = \begin{bmatrix}
\tau_{r} & \tau_{l} 
\end{bmatrix}^T$ where, $\tau_{r}$, $\tau_{l}$ are the torques generated by the right and left wheels respectively.

\subsection{Battery Model}
Each robot in our system is equipped with a Li-Ion battery. The discharge profile of a typical Li-Ion battery with a capacity of $1200 mAh$ is empirically obtained and an analytical model is developed through curve fitting \cite{battery_model}
\begin{equation}\label{discharge_equation}
VI^{n}(D) = \frac{\left(a_{1} + a_{3}D + a_{5}D^{2}\right)}{\left(1 + a_{2}D + a_{4}D^{2} + a_{6}D^{3}\right)}
\end{equation} 
where, variables $a_{1}, a_{2}, \dotsc ,a_{6}$ are coefficients determined by least squares curve fitting, $VI^{n}$ represents the voltage discharge $(D)$, $n$ is a constant ($\approx$ 0.005)\footnote{For detailed battery model, Refer \cite{battery_model}.}.
The voltage during discharge is found using the below equation. 
\begin{equation} \label{eq:26}
V_{t} = \frac{I^{n}V(D)_{t-1}}{I_{t}^{n}} =\left( \frac{I^{n}V(D)_{t-1}}{P_{e}^{n}} \right)^{\frac{1}{(1-n)}}
\end{equation}
where, $I_{t} = \frac{P_{e}}{V_{t}}$, $P_{e}$ is the electrical power, $I_{t}$, $V_{t}$ are the current and voltage at time instant $t$. The discharge capacity for the next discrete time step is the summation of the present battery capacity and the discharge for a $\Delta t$ time interval. \begin{equation}
D_{t} = I_{t} (\Delta t) + D_{t-1}
\end{equation}
% \begin{figure}[h]
% 	\includegraphics[height=5cm,width=8cm, center]{DnVPlot}
% 	\caption{Voltage Vs Discharge Curve for Li-Ion Battery}
% 	\label{DnVPlot}
% \end{figure}
Electrical power $P_{e}$ for each differential drive robot is also given by 

\begin{equation}
P_{e} = \frac{\tau_i \dot{\phi}_i}{\eta}, \ i \in \{\text{left wheel},\text{right wheel}\}
\end{equation}
$\eta$ is robot's motor efficiency. The generated torque per motor $\tau_i$ and the current $I_{w,i}$ per motor is given as,
\begin{align} \label{tor}
\tau_i = K_{t} (I_{w,i} - I_{0}) - b\dot{\phi}_i ,\qquad
 I_{w,i} = \frac{V_i - K_{e}  \dot{\phi}_i}{R}
\end{align} 
$I_{0}$, $K_{t}$, $R$, $K_{e}$, $b$ are the no-load current, torque constant, internal resistance, back-emf constant and damping constant of the DC motor respectively. \par
\begin{figure}[h]
	\includegraphics[scale=0.35, center]{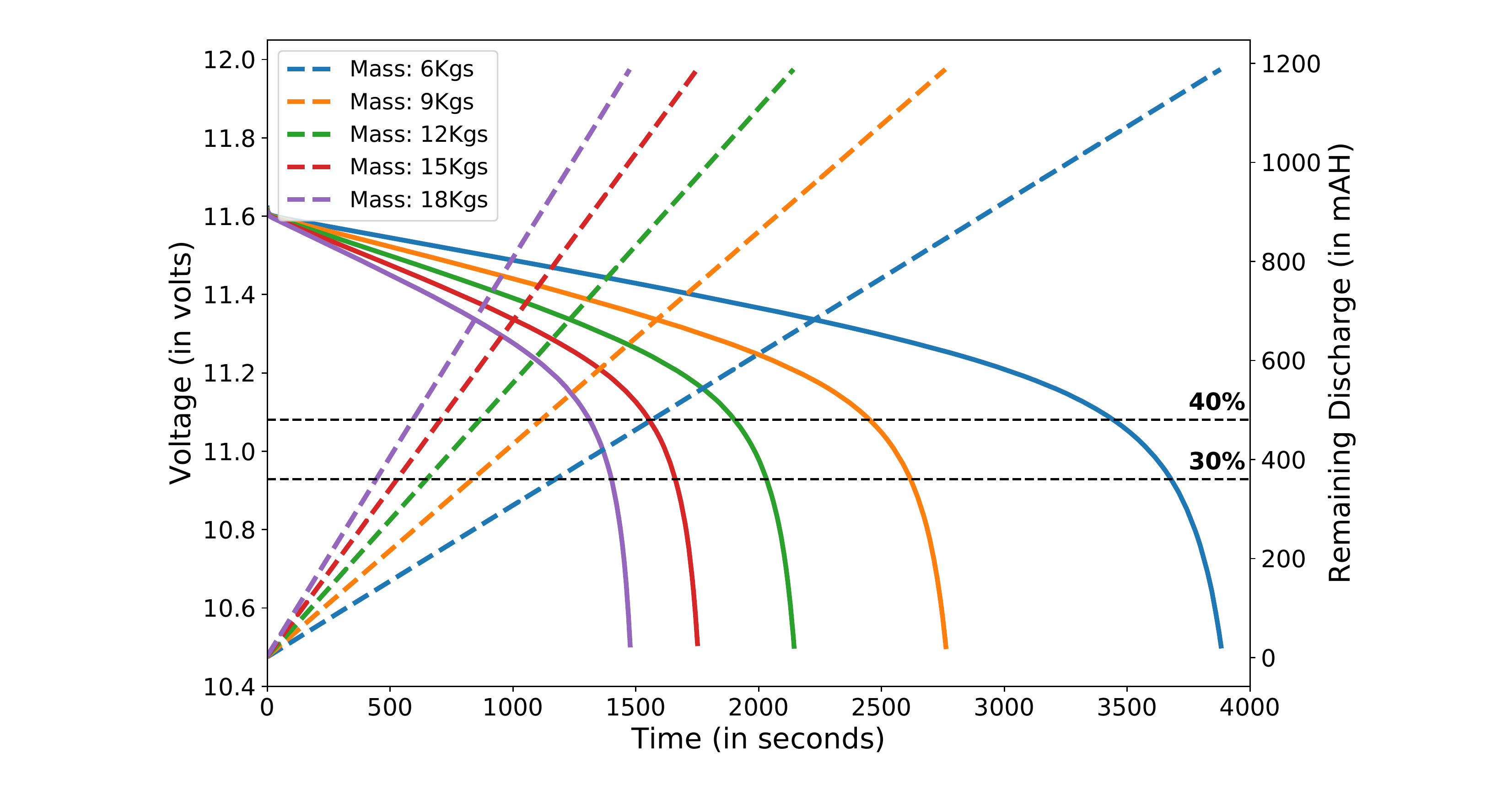}
	\caption{Voltage vs Time plot at different Mass of Payload}
	\label{volt_time_mass}
\end{figure}

Substituting the actuator torque and current equation (\ref{tor}) in dynamics equation (\ref{dynamics}) we obtain the integrated motor, battery and robot dynamics equations which are used %Using the derived integrated dynamics model of a robot, we can estimate the battery discharge duration of each robot in formation. \\
%Using the above battery model, motor model and robot dynamics, we 
to estimate the robot battery discharge where the discharge is directly proportional to the mass. Fig.\ref{volt_time_mass} presents the battery voltage and discharge curves plotted against time for different mass values. The solid lines represent the voltage curves whereas the dotted lines represent the discharge curves. From the figure, it is clear that discharge curves are steeper for heavier payload. The black dotted lines show the discharge at $30\%$ and $40\%$ remaining battery level.

\subsection{Formation Control}
\begin{figure}[h]
	\includegraphics[scale = 0.3, center]{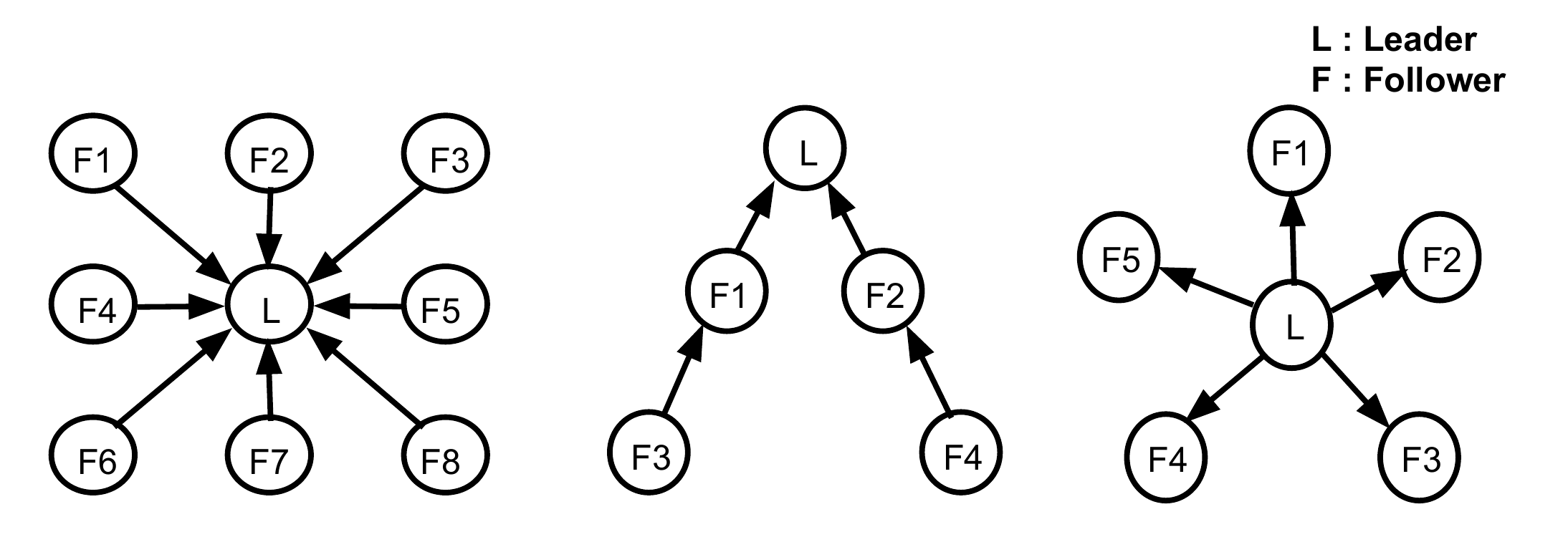}
	\caption{Leader follower formation with different formation shapes}
	\label{formation}
\end{figure}

We implement a decentralized leader-follower formation control law. A leader robot is assigned a predefined closed loop trajectory, generated from a global motion planner \cite{ompl}. All other robots derive their respective pseudo command velocities from the leader's position and command velocities. Fig.\ref{formation} shows the graphical representation of the information flow between the leader and followers. The decentralized control law \cite{formation} is given by the following equations.
\begin{eqnarray} \label{formation_control_eqn}
v_{j} &=& k_{1} \alpha_{j} + v_{i} cos \theta_{ij} - \rho^{d}_{ij} \omega_{i} sin(\psi^{d}_{ij} - \theta_{ij}) \\
\omega_{j} &=& (v_{i}sin\theta_{ij} + \rho^{d}_{ij} \omega_{i} cos(\psi^{d}_{ij} + \theta_{ij}) + k_{2} \beta_{j} + k_{3}\theta_{je})/d \nonumber
\end{eqnarray}
where $\alpha_{j}$ and $\beta_{j}$ is the error in longitudinal and vertical direction respectively which is given by
\small
\begin{eqnarray}
\dot{\alpha_{j}} &=& -k_{4}\alpha_{j} + (k_{5}-\alpha_{j})f_{1j}(x_{je}) - (k_{6} + \alpha_{j})g_{1j}(x_{je}) \\
\dot{\beta_{j}} &=& -k_{4}\beta_{j} + (k_{5}-\beta_{j})f_{2j}(y_{je}) - (k_{6} + \beta_{j})g_{2j}(y_{je}) \nonumber
\end{eqnarray}
\normalsize
and
\begin{eqnarray}
f_{1j}(x_{je}) = Max(k_{1}x_{je}, 0), \quad 
g_{1j}(x_{je}) = Max(-k_{1}x_{je}, 0) \nonumber \\
f_{2j}(y_{je}) = Max(k_{2}y_{je}, 0), \quad 
g_{2j}(y_{je}) = Max(-k_{2}y_{je}, 0) \nonumber
\end{eqnarray}
where constants $k_{1}, k_{2}, k_{3}, k_{4}, k_{5}, k_{6} > 0$, $\rho^{d}_{ij}$ and $\psi^{d}_{ij}$ are the desired distance and orientation to maintain between the leader and follower robot, $v_{i}$ and $\omega_{i}$ are the linear and angular velocities of the leader, $v_{j}$ and $\omega_{j}$ are the generated linear and angular velocities of the $j^{th}$ follower, $d$ is distance from the robot's center to the robot's center of mass. $\theta_{ij}$ is the orientation error of the leader and follower i.e. $\theta_{ij} = \theta{i} - \theta_{j}$ where $\theta_{i}$ and $\theta_{j}$ are the leader and follower orientation respectively. $x_{je}, y_{je}, \theta_{je}$ are the positional tracking errors between the leader and follower.\par
Additionally, we constraint the robots in the formation to maintain a minimum distance $\rho_{ij}^{d}$, such that it guarantees that a path always exits for any robot in the formation to move in and move out of the formation. Therefore, $\rho_{ij}^{d}$ is determined using the dimension of the robot and an approximate error ($\delta$) in the received localization values from the Decawave modules \cite{decawave}.
% \begin{equation}
    % L_{ij} = max(R_{L}, R_{W}) + \sigma + \delta
% \end{equation}
% where $R_{L}$ and $R_{W}$ are the length and width of the robot. $\sigma$ is the dilation length or width of the robot. $\delta$ is the approximate error in the received localization error from the Decawave modules.\par
From the experimental analysis, we analyzed that the value of $\delta$ is nearly equal to $0.05$ meters.
% and $\sigma$ is equal $\frac{1}{4}$ of the robot length or width (whichever is more).

\section{Selection Mechanism for Replacement}\label{Scheduling}
% A mechanism (Sec. \ref{central}) is proposed to identify the robots in the formation which need to be replaced due to low battery charge and consequently find charged robots from the recharge hubs for replacement.
In section \ref{central}, a mechanism is proposed to identify the low battery robots in the formation and consequently find the charged robots from the hubs for replacement. The problem is modeled as an optimization which takes the present battery level of all robots as input and decides on the subset of robots in the formation which have to be replaced (or recharged). When the formation reaches a hub, the optimizer numerically computes a solution to replace the low battery robot with a charged robot present at the current hub and identifies the robots to be replaced. A robot replacement strategy is discussed in section \ref{replacement}. The sequence of steps involved in the proposed approach is shown below.
% Algorithm \ref{alg:algo1} details the steps involved in the proposed approach.
\begin{algorithm}[h]
\SetAlgorithmName{Sequence of Steps}{}
     Select robots to carry a payload \\
	 Make formation based on shape of the payload\\
	 Provide trajectory information to the leader\label{start_again}\\
     Formation starts moving\\
	 Monitor battery status of all the robots\\ 
	 Run Optimization on reaching recharge hubs to get solution vector\\
	 Check if replacement is needed, based on the solution vector\\
     \uIf{$Replacement$}     
        {   
        % \skipnumber 
            % { 
            \uIf{$Replacement \enspace available$}
                { 
                \nonl Piston moves up for all the robots\label{piston_algo}\\
                \nonl Support robot joins the formation without colliding with the neighbour robots \\
                \nonl Piston moves up for the support robot\\
                \nonl Piston moves down for the low battery robot\\
                \nonl Low battery robots moves out of the formation and reaches hub\\
    	        \nonl Charged robot occupies the place vacated by the low battery robot\\
    	        \nonl Piston moves up for the charged robot\\
    	        \nonl Piston moves down for the support robot\\
    	        \nonl Support robot moves out of the formation\\
                \nonl Piston for all the robots in formation moves down\\
                \nonl Go to step $\ref{start_again}$
                } 
            \nonl \Else
                {  %nested else
                \nonl System waits until a replacement is available 
                }
            % } 
        }
    \Else 
    {
        % \skipnumber[1]{
        \nonl Go to step $\ref{start_again}$
        % }
    }
    \caption{Robot Selection and Replacement Process}
    \label{alg:algo1}
\end{algorithm} 

\subsection{Battery Discharge based Selection} \label{central}
In our system, each robot has two states (a) Active state i.e.\ robot is in the formation and carrying the payload (b) Inactive state i.e.\ robot is not in the formation and charging its battery.
A robot in an active state undergoes continuous discharge whereas in inactive state a robot is charged at a hub. A quadratic program is formulated, in order to make decisions which would \begin {enumerate*}[label=\itshape\alph*\upshape)]
\item extend the operating time of the formation,
\item reduce the long-term expected cost of replacing the robots at hubs,
\end{enumerate*}
adhering to system constraints.
% We maximize the operational time of the system by minimizing the battery discharge of the robots in formation. 
% Also, it is difficult to have a same battery level (capacity) in all the robots always at the time of task assignment, we therefore consider different initial battery level of each robot.

\subsubsection*{Optimization Formulation} \label{optimization_formulation}
We model a moving horizon based optimization where we consider a few steps ahead (horizon length) to select a set of robots that contribute to payload transportation.
Let $N$ be the total number of available robots, $k$ be the horizon length ($k=2$ in our work) such that $k<N_{h}$, $N_{h}$ being then total number of recharge hubs. Let $\bm{X}^{j}$ be a $N$ length binary vector represented as $\begin{pmatrix}x_{1}^{j} & x_{2}^{j} & \dotsc & x_{N}^{j} \end{pmatrix}$, $\bm{X}^{j}$ is the solution vector at $j^{th}$ hub, $j = 1, 2, \dotsc, k$ and $x_{i}^{j} \in \{0, 1\}$ for $i = 1, 2, \dotsc, N$ where $x_{i}^{j} = $`$1$' when $i^{th}$ robot is active (discharging) and `$0$' if robot is inactive (charging). $\bm{D}^{j}$ be a $N$ length vector written as $\begin{pmatrix}d_{1}^{j} & d_{2}^{j} & \dotsc & d_{N}^{j} \end{pmatrix}$ which is the discharge vector at $j^{th}$ hub, $d_{min} \leq d_{i}^{j} \leq d_{max}$, $d_{i}^{j}$ corresponds to the battery consumption of the $i^{th}$ robot (in mAh) at the $j^{th}$ hub, where $d_{min}$ means fully charged robot (0 mAh consumed) and $d_{max}$ denotes completely discharged robot (1200 mAh consumed).\par
% A predefined trajectory is considered in this paper for the payload transport system. While going from one recharge hub to other hub, a robot loses its battery by a constant value $r_{d}$, whereas all the other robots are charging and gain battery capacity $r_{c}$ .\\
The estimated battery discharge vector for the robots at $m^{th}$ horizon $(m < k)$ is given by
\begin{align}
\bm{D}^{m} &= \bm{D}^{m-1} + r_{d}\bm{X}^m  + r_{c}(1-\bm{X}^m)
\end{align}
where $r_{c}, r_{d}$ are predetermined charge and discharge constants between any two consecutive hubs respectively.
The more generalized form of the above equation can be written as 
% The solution of the above recurrence relation can be written with respect to the initial state vector ($\bm{X}^0$) and initial discharge vector ($\bm{D}^0$) which is given below
\begin{align}\label{discharge_eqn}
\bm{D}^{m} &= \bm{D}^{0} + m r_{c} + (r_{d} - r_{c}) \displaystyle\sum_{j=1}^{m} \bm{X}^{j} 
\end{align} where $\bm{D}^{0}$ is the initial discharge vector.
% \begin{equation*}
% \bm{D}_{1} = \bm{D}_{0} + r_{d} \bm{X}_{0} + r_{c} (\bm{1}-\bm{X}_{0})
% \end{equation*}
% where $\bm{X_{j}}$ and $\bm{D_{j}}$ represents the robot vector states and their discharge values at the $j^{th}$ hub (`$0$' representing the current hub), where j = $0, 1, 2, \dotsc, k$. Rearranging the above equation, we get
% \begin{align*}
% \bm{D}_{1} &= \bm{D}_{0} + r_{c} + (r_{d} - r_{c}) (\bm{X}_{0})
% \end{align*}
% For horizon length ($k=2$),
% \begin{align*}
% \bm{D}_{2} &= \bm{D}_{1} + r_{d} \bm{X}_{1} + r_{c} (\bm{1}-\bm{X}_{1}) \\
% \bm{D}_{2} &= \bm{D}_{0} + 2r_{c} + (r_{d} - r_{c}) (\bm{X}_{0} + \bm{X}_{1})
% \end{align*}
% Similarly, for $k$ horizons, 
% \begin{align}\label{discharge_eqn}
% \bm{D}_{k} &= \bm{D}_{0} + k r_{c} + (r_{d} - r_{c}) \displaystyle\sum_{i=1}^{k} \bm{X}_{i} 
% \end{align}
We minimize the battery discharge of the robots in formation at the $k^{th}$ hub, to ensure longer operational time of the complete system.  Using Equation(\ref{discharge_eqn}), we can write the discharge of active robots at $k^{th}$ hub as $(\bm{D}^{k})^{T} \bm{X}^{k}$.\par
Note, higher the number of replacements made at any hub point, the more time system remains idle, causing delay in the transportation. Therefore, we also maximize the robot retention at any recharge hub, i.e. to keep the same robots in formation, which ensures carrying out minimum number of replacements at a hub point. Let $\bm{H}^{j}$ be a $N$ length binary vector represented as $\begin{pmatrix}h_{1}^{j} & h_{2}^{j} & \dotsc & h_{N}^{j} \end{pmatrix}$ for $j^{th}$ hub, $h_{i} \in \{0, 1\}$ for $i = 1, 2, \dotsc, N$ and $j = 0, 1, \dotsc, N_{h}-1$ . $h_{i}^{j} = $`$1$' if $i^{th}$ robot is present at the $j^{th}$ hub and `$0$' otherwise. Here, retention refers to using a robot for two consecutive state vectors (hubs). Hence, the retention of robots ($\bm{R}_{r}$) for $k$ horizons is written as
% \squeezeup
\begin{align}
\bm{R}_{r} = \displaystyle\sum_{j=1}^{k} \bm{X}^{j}\bm{X}^{j-1}
% rr = \bm{X}_{0}\bm{X}_{1} + \bm{X}_{1}\bm{X}_{2} + \cdots + \bm{X}_{k-1}\bm{X}_{k}
\end{align}
The optimization can now be modeled as a quadratic program as shown below:
\begin{align} \label{objective}
minimize \quad \frac{1}{2} \bm{X}^{T}\bm{PX} + \bm{Q}^{T}\bm{X}
\end{align}
\begin{align*}
subject \enspace to: \quad & Formation \enspace Constraint \enspace (\ref{fconst})\\
&Battery \enspace Constraint \enspace (\ref{bconst})\\
&Hub \enspace Constraint \enspace (\ref{hconst})
\end{align*}
where $\bm{X}$ = $\begin{pmatrix}\bm{X}^{0} & \bm{X}^{1} & \dotsc & \bm{X}^{k} \end{pmatrix}^{T}$ is $1 \times kN$ length optimization vector. $\bm{P}$ is a $kN \times kN$-dimension real symmetric matrix, $\bm{Q}$ is a $kN \times 1$-dimension real valued vector. 
\begin{align*}
\bm{P} &= w_{1} \bm{P}_{d} - w_{2} \bm{P}_{r} \\
\bm{Q} &= w_{1} \bm{Q}_{d} - w_{2} \bm{Q}_{r}
\end{align*}
$\bm{P}_{r}, \bm{P}_{d}$ are the Quadratic terms in the objective denoting the coefficients of robot retention and battery discharge respectively. Here $\otimes$ refers to the Kronecker product.\par
\begin{align*}
\bm{P}_{r} =
 \begin{bmatrix}
  0 & 1 & 0 & \cdots & 0 & 0 \\
  1 & 0 & 1 & \cdots & 0 & 0 \\
  0 & 1 & 0 & \cdots & 0 & 0 \\
  \vdots  & \vdots & \vdots  & \ddots & \vdots & \vdots \\
  0 & 0 & 0 & \cdots & 0 & 1\\ 
  0 & 0 & 0 & \cdots & 1 & 0 
 \end{bmatrix}_{k \times k} \otimes \bm{I}_{N\times N} , \quad
\bm{P}_{d} =
 \begin{bmatrix}
  0 & 0 & \cdots & 1 \\
  0 & 0 & \cdots & 1 \\
  \vdots & \vdots  & \ddots & \vdots \\
  1 & 1 & \cdots & 2 
 \end{bmatrix}_{k \times k} \otimes \bm{I}_{N\times N}
\end{align*}
$\bm{Q}_{r}, \bm{Q}_{d}$ are the linear terms of the objective referring to the coefficients of robot retention and battery discharge respectively. 
\begin{align*}
\bm{Q}_{r} &= \begin{bmatrix}
  1 & 0 & 0 & \cdots & 0 & 0 \\
  \end{bmatrix}_{1 \times k}^{T}  \bm{X}^{0} ,
\end{align*}
\begin{align*}
\bm{Q}_{d} &= \begin{bmatrix}
  0 & 0 & 0 & \cdots & 0 & 1 \\
  \end{bmatrix}_{1 \times k}^{T} (\bm{D}^{0} + k r_{c})
\end{align*}
Equation (\ref{objective}) is therefore a result of weighted sum of two objectives where $w_{1}, w_{2}$ are the weights given to each objective.\par
The \textbf{formation constraint} limits the number of robots which can participate in payload transportation, given as
\begin{align} \label{fconst}
\bm{C} \bm{X}^{j} = F \quad \forall  j = 0, 1, \dotsc, k
\end{align}
where $F$ is a scalar, denoting number of robots in formation, $\bm{C} = \bm{1}_{1\times N}$
% \begin{align*}
% C = I_{k \times k} \otimes \bm{1}_{1 \times N}
% \end{align*}
This will ensure that only $F$ number of robots are present in the formation.\par
The \textbf{battery constraint} ensures that at each hub, the battery of each robot in the formation is above a certain threshold capacity ($d_{th}$). Using Equation (\ref{discharge_eqn}), we can model the battery constraint as
 \begin{align}\label{bconst}
    d_{i} x_{i} \le d_{th}  \quad \forall  i = 0, 1, \dotsc, N
 \end{align}\par
The \textbf{hub constraint} helps in choosing only those robots for the replacements which are available at the current hub. At any horizon, the number of robots retained and the number of robots replaced at the current hub, must sum to the number of robots in a formation. This can be written as 
\begin{align}  \label{hconst}
(\bm{X}^{j-1} + \bm{H}^{j})^{T} \bm{X}^{j} = F \quad \forall  j = 1, 2, \dotsc, k
\end{align} 

The quadratic optimization is solved using CPLEX solver. The optimization takes the robots' battery parameters (measured voltage and current values) as input and generates a binary solution vector. The solution vector generated at the current hub and the previous hub are compared to find the robots to be replaced. The index of each element of the solution vector corresponds to robot id. For e.g.\ considering ten robots in total and three robots in formation, let the solution vector at the previous hub be $[\begin{matrix} 1 & 0 & 0 & 1 & 0 & 1 & 0 & 0 & 0 & 0  \end{matrix}]$ which represents the robots with id \{$1, 4, 6$\} were in formation before reaching the current hub. And let the solution vector at current hub after optimization be $[\begin{matrix} 0 & 1 & 0 & 1 & 0 & 1 & 0 & 0 & 0 & 0  \end{matrix}]$ which represents the robots with id \{$2, 4, 6$\} should be in formation. We therefore see that the robot with id $1$ (low battery robot) needs replacement with robot with id $2$ (charged robot).

% \begin{align*}
% \bm{Q}_{r} &= \begin{bmatrix}
%   1 & 0 & 0 & \cdots & 0 & 0 \\
%   \end{bmatrix}_{1 \times k}^{T} \otimes \bm{X}^{0} ,
% \end{align*}

\subsection{Robot Replacement}\label{replacement}
In the previous subsection, we discussed the optimization formulation in detail. Numerically evaluating the quadratic integer program, identifies the robots in a formation which have to be replaced with charged robots at a recharge hub. We need to make sure that the payload on top of the robots must not be affected during robot replacement. To facilitate robot replacements, a piston is mounted on each robot to move the payload up and down in case of any failure.
% \begin{figure}[h]
% 		\captionsetup[subfigure]{font=scriptsize,labelfont=scriptsize}
%         \hspace{5mm}
%         \begin{subfigure}[b]{0.18\textwidth}
%                 \includegraphics[width=2.8cm, height=2.8cm]{p2}
%                 \caption{Piston on the robot (Normal Position)}
%                 \label{piston}
%         \end{subfigure}%
%         \hspace{8mm}
%         \begin{subfigure}[b]{0.18\textwidth}
%                 \includegraphics[width=2.8cm, height=2.8cm]{p1}
%                 \caption{Piston on the robot (Stretched Position)}
%                 \label{piston1}
%         \end{subfigure}%
%         \caption{Piston Design in UP and DOWN position}\label{replacement_piston}
% \end{figure}
% \begin{figure}[h]
% 	\includegraphics[height=2cm,keepaspectratio, center]{piston}
% 	\caption{A piston with translational motion installed on the robot}
% 	\label{piston}
% \end{figure} 
When a  battery failure occurs, the piston on all the robots moves up to push the payload such that a new robot can join the formation. To ensure payload stability, support robots are utilized to temporarily balance the system until the robots are replaced. The low battery robot moves its piston down and leaves the formation to recharge its battery at the hub. A charged robot (solution provided by the optimization) from the hub reaches the formation and occupies the space vacated by the low battery robot. The piston of the charged robot moves up, such that the support robot can now leave the formation. The support robot lowers down its piston and moves out of the formation. And hence the formation starts traversing its trajectory again. This completes the replacement process.\par
% of replacing the low battery robot with a charged robot. \par
Collision-free path planning of robots from recharge hub to the formation and vice-versa is done using a global planner \cite{ompl}. The inter-robot distance is constrained to be greater than the dimensions of a single robot to facilitate robot navigation and replacements. We focus on failure handling of both triangular and rectangular shaped formation for now, but this can be extended to a formation of different shapes. The system can handle any number of replacements at a hub if charged robots are available for replacement at this hub.  Also, as the paper deals with a decentralized leader-follower based formation control \cite{formation}, the replacement criteria remain the same for the leader and followers. When a leader robot loses its battery charge, the id of the new robot (charged robot) is shared with all the followers. The followers then start following the new robot and hence the formation continues. If there is no robot at the hub having sufficient battery for making any replacement or the number of replacement to be carried out is more than the number of robots present at the hub, the optimization fails to compute a solution and the system operation is halted until a replacement is available. In this paper, the idea of robot replacement is carried out to maximize the operating time of the system, however, this replacement mechanism is also applicable in case of other failures in a robot if the localization and mobile capabilities of a robot are not compromised.
\begin{table}[h]
\centering
\captionsetup{ width= 65mm}
\captionof{table}{List of Parameters for each robot ($a_{1} - a_{6}$ are curve fitting gains for battery and $k_{1} - k_{6}$ are control gains for formation of robots)}
\begin{tabular}{ |M{4cm}||M{2cm}|  }
 \hline
 Parameter Name & Value\\
 \hline
 Wheel radius ($r$)(in meters)    &0.035\\
 Wheel base($l$)(in meters)&0.115\\
 $a_{1}$ & 12\\
 $a_{2}$ & 3.409\\
 $a_{3}$ & 39.55\\
 $a_{4}$ & -0.002653\\
 $a_{5}$ & -0.03203\\
 $a_{6}$ & -8.112$e^{-8}$\\
 $k_{1}$ & 1.5\\
 $k_{2}$ & 1.0\\
 $k_{3}$ & 0.025\\
 $k_{4}$ & 15.0\\
 $k_{5}$ & 1.0\\
 $k_{6}$ & 1.0\\
 Torque Const. ($K_{t}$)(in Nm/A)& $28.24 \times 10^{-3}$ \\
 Armature resistance(R)(ohms)&2.4\\
 Chassis weight(m)(in Kgs)& 1.5\\
 Wheel weight($m_{w}$)(in Kgs)& 0.1\\
 Payload weight(in Kgs)& 1-18\\
 \hline
\end{tabular}
\label{param}
\end{table}

\section{Results} \label{simulations}
% In this section, we show the result of above formulated problem by running both simulations and physical hardware on a group of differential drive robots.
We present simulation and hardware results in this section.
Table \ref{param} is a reference to some important parameters and its values, used in the paper.
\subsection{Simulation Results}
 The simulation is performed using twelve robots and three recharge hubs. A formation of three robots is continuously traversing a circular trajectory to show the viability of the solution. In the uploaded video\footnote{Video at: https://youtu.be/-6ivGT3dOQw} we showcased additional hardware results with more number of robots. The average distance of the recharge hubs from the trajectory is around $0.5$ meters.
 \begin{figure}[h]
    \centering
	\includegraphics[width=\textwidth, height=7cm, center]{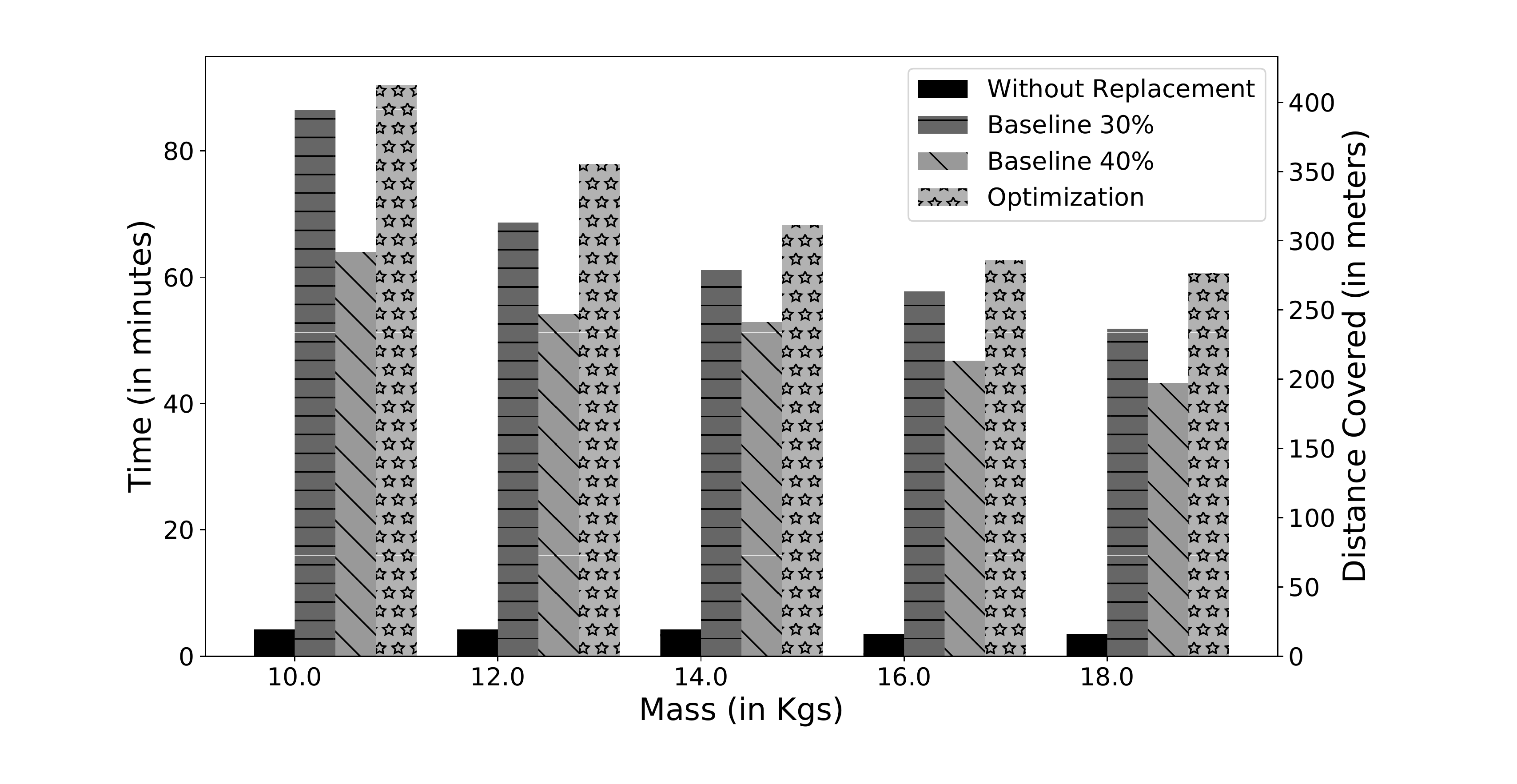}
	\caption{Payload Operating Time Comparison}
	\label{mvtt}
\end{figure}
The trajectory shape is not limited to a circular path but any random trajectory can be considered to keep the notion of recharge hubs and replacements intact. Three out of twelve robots are used in the formation, while others are distributed over the recharge hubs (excluding the support robots). We introduce and compare our proposed approach to a naive baseline approach. The two approaches include (a) baseline approach, where the robot in the formation get replaced as soon as it loses its battery below a threshold voltage. (b) The second approach is our optimization based approach where we maximize the battery state of robots in the formation for $k^{th}$ horizon in advance. The simulations are performed at $k=2$. The choice of k depends on inter-hub distance. If the distance between hubs is large, the value of k is small typically less than 3. We observed that a high value of $k$ would violate the constraints in the optimization as some of the robots are completely discharged before reaching the $k^{th}$ hub point.\par

% introduce and compare our proposed approach to a naive baseline approach.
% Larger value of $k$ will be more time consuming in terms of generating optimized solution. The choice of $k$ depends on how far the hubs are placed, if the distance between the two consecutive hubs is large, a smaller $k$ is preferable as large $k$ in this case will not allow the battery constraints to satisfy, resulting in ending the optimization abruptly.  \par
% If the robot battery is insufficient to reach the $k^{th}$ hub, it is replaced. Because of this predicting nature, we get a increased battery life of the system with the optimization based approach. \par
%  \begin{figure*}[h]
% 	\includegraphics[height=7cm,width=\textwidth, center]{fb}
% 	\caption{Battery Level of robots in formation Vs time (Mass 5 Kg)}
% 	\label{fb}
% \end{figure*}
Fig. \ref{mvtt} compares the battery life of the two approaches for different weights with a system having no replacement strategy. The robots move in a formation carrying different payload mass. Varying the mass affects robot dynamics which draws a high discharge current and hence increases the battery discharge for heavier mass and vice-versa. Fig. \ref{mvtt} shows that as the mass on the system increases, the battery life of the system reduces. The active time of the system without replacement suffers an early breakdown as there are no robots for replacement. As all the experiments are performed using the same seed value for the initial battery charge (without replacement, baseline, optimization), therefore the distance traveled by the formation without replacement is nearly the same. The system stops as soon as any robots run out of battery. Both baseline and optimization approaches work better than the system without replacement. 
\begin{figure}[htb]
    \hspace{-8mm}
    \centering
	\includegraphics[width=\textwidth, height=7cm, center]{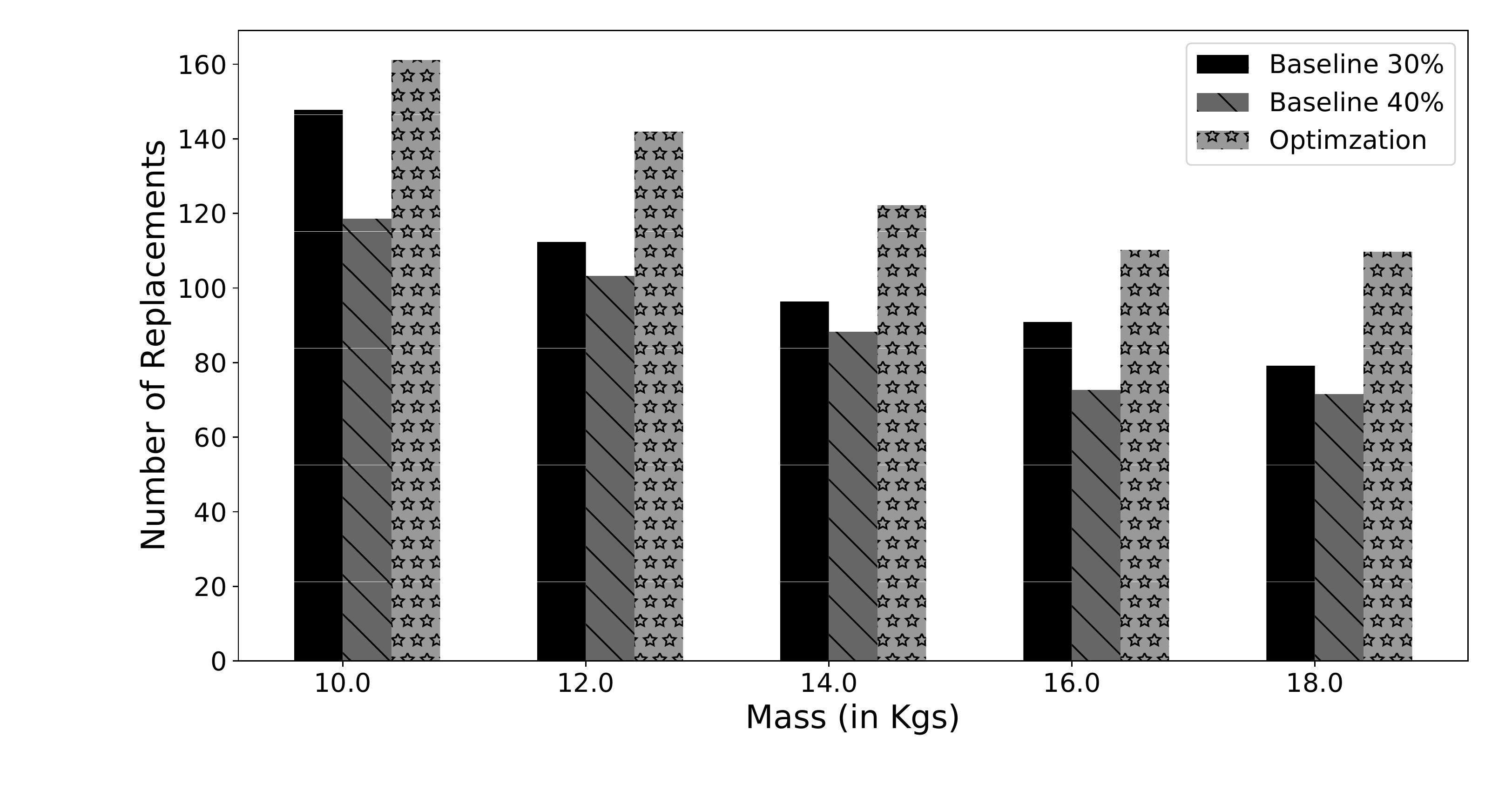}
	\caption{Total Robot Replacement counts}
	\label{mvrr}
\end{figure}
Also, it can be observed that the baseline approach with threshold level $30\%$  travels more distance than the with $40\%$ battery threshold. Similarly, if the threshold is increased to $60\%$, the distance traveled by the formation will be even lesser with more number of replacement counts. The optimization on the other hand (the rightmost bar), provides much improved result in terms of increased battery life with a minimal increase in the number of replacements.\par
% \begin{figure*}[ht]
% 	\includegraphics[width=14cm, height=5.5cm,center]{batt_prof}
% 	\caption{Battery profile for the robots in formation}
% 	\label{Batt_prof}
% \end{figure*}
%  \begin{figure*}[h]
% 	\includegraphics[width=16cm, height=6
% 	.cm, center]{avg_dis}
% 	\caption{Replacement Vs Average Battery Discharge}
% 	\label{hist}
% \end{figure*}\par
\begin{figure*}[h]
    \centering
	\includegraphics[width=\textwidth, height=6.5cm,center]{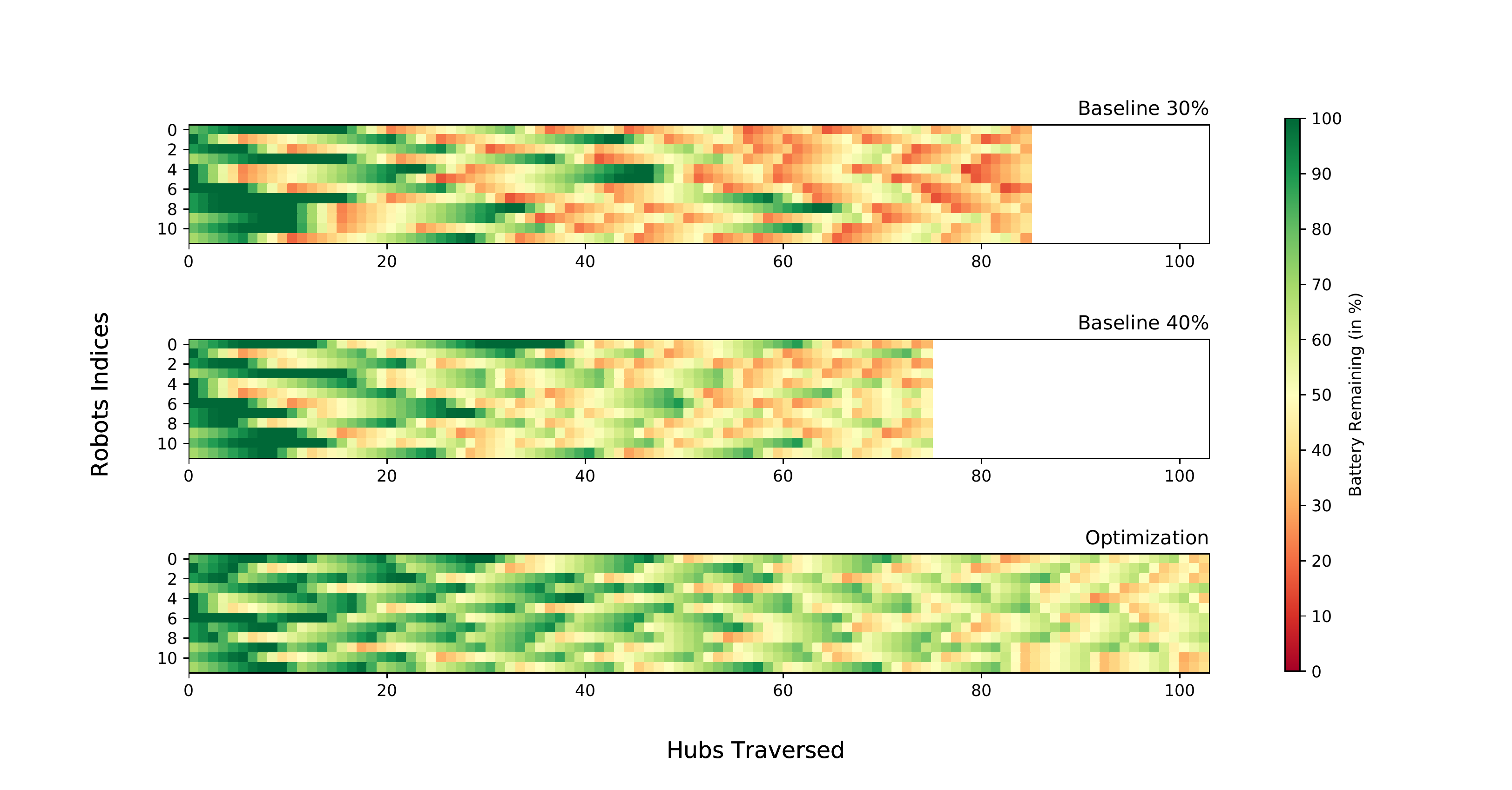}
	\caption{Battery profile for the robots in formation}
	\label{Batt_prof}
    % \vspace{-5mm}
\end{figure*}
\begin{figure*}[htb]
    \centering
	\includegraphics[width=16cm, height=6.5cm, center]{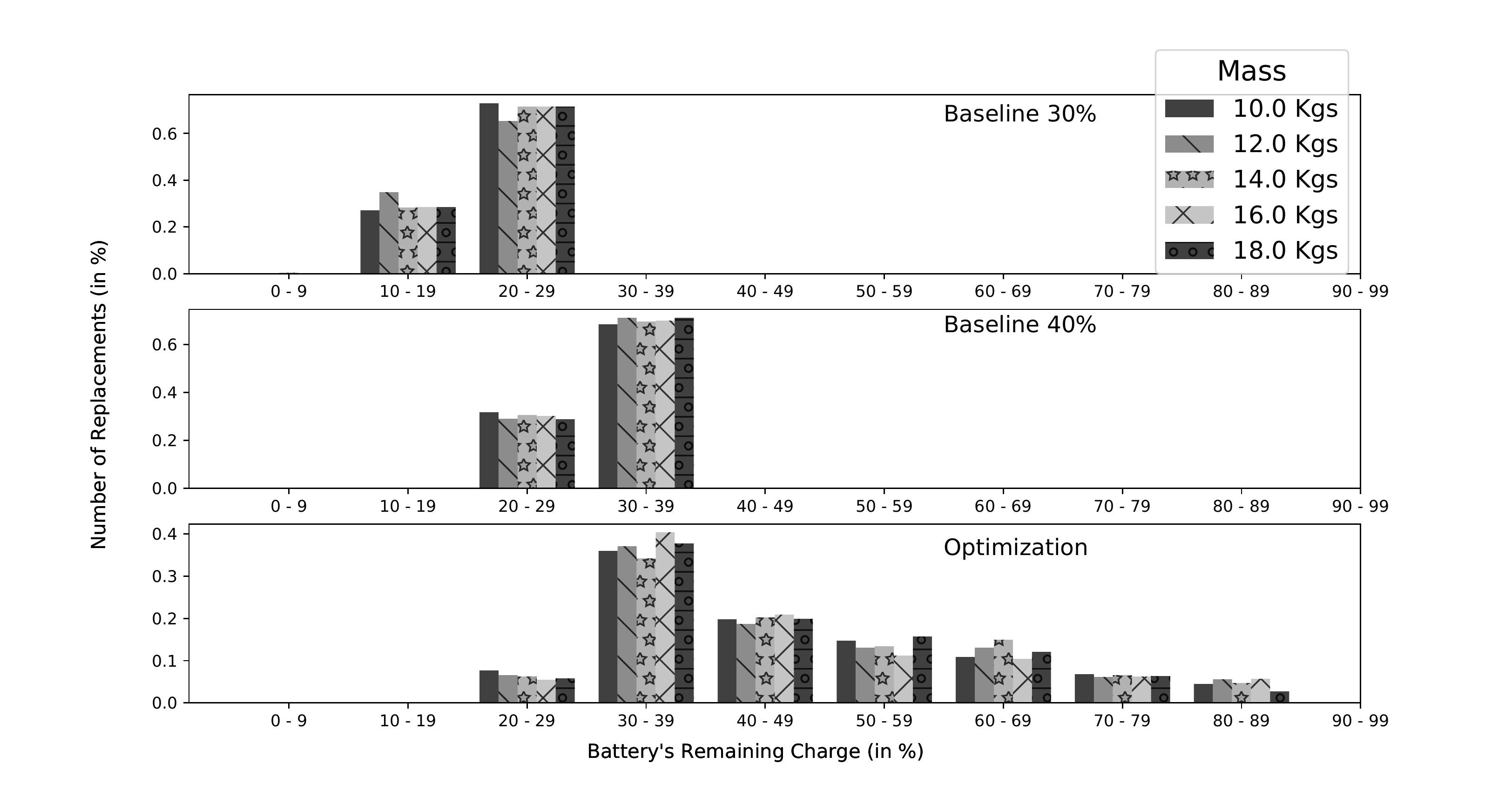}
	\caption{Replacement Vs Average Battery Discharge}
	\label{hist}
\end{figure*}\par
Fig. \ref{mvrr} highlights the number of replacements through the lifetime of the system. Each count refers to a robot being replaced at a recharge hub. It is intuitive to think that the robots following the baseline approach ($30\%$ threshold) will experience less number of replacements than with ($40\%$ threshold), as a robot moves out of formation only if the battery level is below the defined threshold. In the optimized approach, though the robot replacement count is slightly higher, it allows the formation to travel a longer distance (ensuring longer operating time). The replacements are made such that the \textit{Travel time} is more and the total \textit{Replacement Time} is less. Here, \textit{Travel time} corresponds to the time utilized in transporting the payload and \textit{Replacement time} is the time taken for replacing the low battery robot with a new one. A balance is maintained between the recharging time and the active time of the battery. \par
Fig. \ref{Batt_prof} displays the battery discharge profile for all the robots vs the total hub traversed by the system. \textit{Each robot is displayed with respect to its remaining charge where the dark green shade represents a charged robot that turns lighter as robot discharges. The color turns red if the robot is fully discharged}. It can be seen from the plot that the battery level of the robots (having the same initial battery) is maintained at a higher value in case of optimization than the baseline approach. Hence, the figure shows that the system lasts longer with optimization approach.
% The figure also shows that the robots in baseline approach are drained out of battery much earlier than the robots in the optimization approach. This makes the system run to last longer than the other two approaches.   
% Fig. \ref{fb} displays the plots denoting the battery level of each robot in formation through out the lifetime of the system. The three subplots refer to the battery level of its corresponding robot in formation. The blue lines indicates the baseline approach at $20\%$ threshold and orange line denotes the baseline approach with $40\%$ battery threshold where the robots experience a continuous fall in their battery level and the replacement occurs whenever a the robot in formation goes beyond the threshold. Whereas the effect of optimization can be seen in the curves marked with yellow color, where the robots in formation are replaced before going very low in charge. This is because of the predicting nature of the optimization which allows the robot to replace if it cannot reach the next recharge hub. This results in increase battery life of the system and also allow the robots to recharge quickly.\\

%  \begin{figure*}[h]
% 	\includegraphics[height=7cm,width=\textwidth, center]{rcb}
% 	\caption{Replacements Vs battery discharge range comparison}
% 	\label{rcb}
% \end{figure*}
One key aspect of the optimization is to note the range of battery's remaining charge when the replacement occurs. Fig.\ref{hist} shows a comparison of replacement counts with the battery level range for baseline and optimization approach. The baseline approach replaces the robot when the robot battery voltage is less than the threshold battery voltage. The replacement window for $30\%$ battery threshold ranges from $10-30\%$ of the total battery charge. This is because of the fact that the robots do not leave the formation even when the battery voltage level is just above the threshold (say $31\%$), which led the replacements to occur in a threshold window lower than the set threshold. Similarly, with $40\%$ threshold, the replacement window lies in $20-40\%$ of total battery range.   
% When the threshold is kept at $30\%$, the replacement occurs when battery ranges in $(0-30)\%$ while at threshold $40\%$, this range shifts to $(0-40\%)$. 
\textit{However the optimization approach, in comparison to the baseline approach, replaces the robot much before the lowest battery level which allows a robot to recharge itself quickly and thus helps in reducing the charging time of the robot. In optimization, the robot gets replaced even when the battery's remaining charge is in the range of $80-90\%$. As we have modeled our optimization considering future time steps, it sometimes replaces a charged robot at a particular hub so as to avoid constraint violations, due to the absence of charged robot on the hub}.\par 
% . For e.g.\ Assume that the robots in formation have sufficient battery such that no replacement is needed at the present hub. Also, there are no recharge robots available at the present hub. In such a scenario, when the formation will reach the same recharge hub after traversing the complete trajectory once, there might not be any robot available for replacement then. 

% This will ensure that few robots are available a the recharge hubs to carry out replacement. 
% to make sure that each robot in the formation still has enough battery to reach the next recharge hub. The optimization does not allow a robot to reach very low battery level and thus helps in reducing the charging time of the robot.\par
% The battery range of replacement for optimization based approach is near $(30-60)\%$. 
% Also, at heavier payloads the discharge rate is higher and hence the optimizer replaces robots at a higher battery level to ensure that the robots in formation travel more distance.\par
Thus, the optimization increases the operational time of the system in comparison to the baseline approach.
% , while having replacement robots almost always.

\subsection{Implementation on Robots}  \label{hardware_implementation}
The hardware implementation\footnote{Video at: https://youtu.be/-6ivGT3dOQw} is carried out on a custom built non-holonomic differential drive robots, each having a Raspberry Pi and an Arduino Uno board mounted on it. The robot localization is achieved by fusing the sensor data of Ultra Wide Band transceiver \cite{decawave} (Decawave DWM1000), robot's wheel odometry and an Inertial measurement unit(IMU-MPU9250), giving a positional accuracy of $<6$ cms. We use an extended Kalman filter to fuse information from decawave modules, imu, and wheel odometry to localize the robot.
% We use robot localization \cite{robot_localization_ros} and imu complementary filter \cite{imu_complementary_filter}, which are built in ROS (Robotics Operating System) packages for using Extending Kalman Filter (EKF) on positional coordinates and orientation data, received from odometry, decawave and IMU sensor.  
A linear actuator(Piston) is mounted on all the robots to lift the payload up and down to facilitate robot replacement. Each robot can carry a payload weight of around seven kg. All the robots share their information wirelessly on a locally created wifi network. Raspberry Pi does the high level processing such as robot coordination, path planning, etc, with a ROS (Robot Operating System) supported environment. Arduino Uno is used to perform a low level control to execute the command velocities on the robot. It runs PID (Proportional Integral Derivative) control loop to maintain desired wheel velocities of a robot with a loop frequency of 20 Hz. The total cost of each robot is around $\$ 250$. PYTHON is used as the programming language to generate the hardware and simulation results.\par
\begin{table}
\centering
\captionof{table}{Follower desired Distance($\rho^{d}_{ij}$) (meters) and Angle($\psi^{d}_{ij}$) (degrees) from the leader}
\begin{tabular}{ |M{2cm}|M{3cm}|M{3cm}|c| } 
\hline
Follower & Parameter & Value \\
\hline
\multirow{2}{4em}{1} & $\rho^{d}_{ij}$ & 0.6 \\ 
& $\psi^{d}_{ij}$ & 0 \\ 
\hline
\multirow{2}{4em}{2} & $\rho^{d}_{ij}$ & 0.6 \\ 
& $\psi^{d}_{ij}$ & 90 \\ 
\hline
\multirow{2}{4em}{3} & $\rho^{d}_{ij}$ & 0.6 \\ 
& $\psi^{d}_{ij}$ & 180 \\ 
\hline
\multirow{2}{4em}{4} & $\rho^{d}_{ij}$ & 0.6 \\ 
& $\psi^{d}_{ij}$ & -90 \\ 
\hline
\end{tabular}
\label{robot_dist_form}
\end{table}
A total of eight robots were used to perform physical validation with two recharge hubs present at the periphery of the trajectory. Five robots are moving in a formation with random initial battery levels and two robots are kept at recharge hubs with nearly full batteries. One support robot is also present on the periphery of the trajectory. The robots were kept in a square formation (four robots at the corners and one in the center). Table \ref{robot_dist_form} contains the values of the desired distance and angle to be maintained by all the followers from the leader. The linear and angular velocity of the robot is $6 \enspace cms/sec$ and $0.05 \enspace radians/sec$. Each robot's battery is monitored and shared with the central server which decides on which robot needs to be replaced on the recharge hub (using optimization). CPLEX optimizer by IBM \cite{cplex} is used to run the optimization. Fig. \ref{hardware} shows the experimental setup including replacements. After running multiple experiments, we have observed that it takes about 3 minutes for a robot to get replaced at a recharge hub. The complete experimentation has also been implemented and validated on Gazebo simulator. 
\begin{figure}[ht]
\centering
% \subfigure[Topview for Gazebo simulator environment]{\includegraphics[height=5.7cm,width=.5\linewidth]{map.pdf}}
% \hfill
\includegraphics[height=5.7cm,width=\textwidth, center]{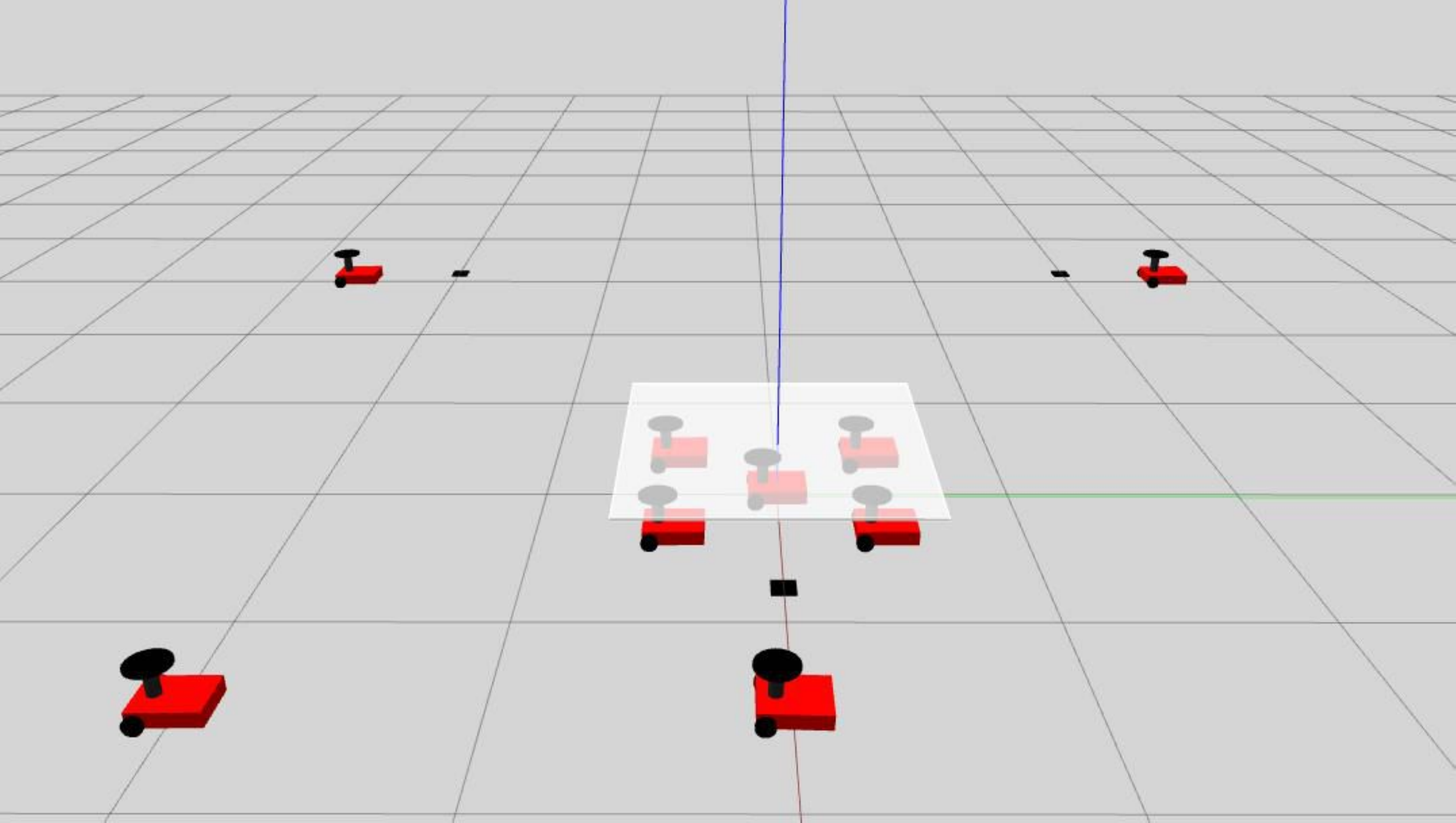}
% \subfigure[Sideview for Gazebo simulator environment]{\includegraphics[height=5.7cm,width=\textwidth]{map2.pdf}}
\caption{Experimental Initial gazebo setup: Nine robots are considered in total. Five robots in a formation are carrying a payload. Three recharge hubs, each containing one charged robot, are present along the trajectory. A support robot is also present along the trajectory to assist robot replacement.}
\label{gazebo}
\end{figure}

% \begin{figure}
% 	\includegraphics[scale=0.4, center]{gazebo}
% 	\caption{Experimental Initial gazebo setup}
% 	\label{gazebo}
% \end{figure}
Initial set up of the gazebo simulator is shown in Fig.\ref{gazebo}, where robots move in a circular trajectory containing three hubs. We showcase\footnote{Video at: https://youtu.be/-6ivGT3dOQw} that the support robot is not always used while replacements are carried out.

\section{Conclusion} \label{conclusion}
We show the validity of a loosely coupled payload transport system with robot replacement. We built our custom differential drive robots having payload lifting capabilities. The robots carry a payload from one place to another while moving in a formation and handles any low battery failure in the system to extend the operating time of the system. We presented an algorithm for task constrained robot replacement to increase the operational duration of a multi-robot payload transport system. We formulated an integer quadratic program to identify the low battery robots to be replaced with charged robots to ensure that the system remains operational. The charged robots are present at the recharge hubs that are located on the periphery of the trajectory. Support robots are used in critical replacements where there are chances of unbalanced payloads. We showcased the results of our approach through extensive simulation results and hardware validation results for robot replacement within a formation. Various test cases are considered including multiple replacements at a single hub and multiple replacements at multiple hubs. Formations with four robots and five robots carrying payload are showcased undergoing replacements in case of low battery failure.\par
Future work would involve decentralizing the scheduling algorithms to enable a distributed multi-agent dynamic task allocation. The concept of stationary recharge hubs can be replaced by moving recharge hubs. The present work includes a few formation shapes, which can be extended to different formation shapes. Replacing the robots in moving formation can be explored.

\section{Acknowledgement} \label{ack}
The author would like to thank Shrey Agrawal, Ayush Gaud for their help with the physical validation.
% In this work, we presented an algorithm for task constrained robot replacement to increase the  operational duration of a multi-robot payload transport system. We formulated an integer quadratic program to identify robots to be replaced with charged robots to ensure that the system remains operational. Recharging stations (Hubs) are placed at the periphery of the trajectory to make replacements of the low battery charged robots.  We showcased the results of our approach through extensive simulation results and hardware validation results for robot replacement within a formation. Future work would involve decentralizing the scheduling algorithms to enable a distributed multi-agent dynamic task allocation. The stationary recharge hubs will be made moving hubs, which can change their position based on the fault site, where the battery failure is occurring. The present work only includes few formation shapes, which will be extended to other different shaped formation. It will be a huge challenge for us to make the robot replacement without even stopping the formation at the recharge hubs in case of any failure.

\bibliographystyle{splncs04}
\bibliography{root}

\end{document}